\documentclass{article} 
\usepackage{iclr2022_conference,times}

\usepackage{amsmath,amsfonts,bm}

\def\eqref#1{equation~\ref{#1}}

\def\1{\bm{1}}

\def\rvc{{\mathbf{c}}}
\def\rvd{{\mathbf{d}}}

\def\rvh{{\mathbf{h}}}

\def\rvm{{\mathbf{m}}}

\def\rvx{{\mathbf{x}}}

\def\vc{{\bm{c}}}

\DeclareMathAlphabet{\mathsfit}{\encodingdefault}{\sfdefault}{m}{sl}
\SetMathAlphabet{\mathsfit}{bold}{\encodingdefault}{\sfdefault}{bx}{n}

\def\gC{{\mathcal{C}}}

\def\gE{{\mathcal{E}}}
\def\gF{{\mathcal{F}}}
\def\gG{{\mathcal{G}}}

\def\gL{{\mathcal{L}}}

\def\gN{{\mathcal{N}}}

\def\gV{{\mathcal{V}}}

\def\gX{{\mathcal{X}}}
\def\gY{{\mathcal{Y}}}

\def\sR{{\mathbb{R}}}

\usepackage{amsmath, bm, amssymb, amsthm}
\usepackage{multirow}
\usepackage{mathtools}
\usepackage{threeparttable}
\usepackage{tablefootnote}
\usepackage{xspace}
\usepackage{graphicx}
\usepackage{subfigure}

\usepackage[utf8]{inputenc} 
\usepackage[T1]{fontenc}    
\usepackage[colorlinks,
    linkcolor=blue,
    anchorcolor=blue, 
    citecolor=black,]{hyperref}       
\usepackage{cleveref}
\usepackage{url}            
\usepackage{booktabs}       
\usepackage{amsfonts}       
\usepackage{nicefrac}       
\usepackage{microtype}      
\usepackage{xcolor}         
\usepackage{algorithm}
\usepackage{algorithmic}
\usepackage{wrapfig,tikz}
\usepackage{chemformula}
\usepackage{footmisc}
\usepackage{lipsum}

\usepackage{siunitx}

\usepackage{tikz}
\usepackage{pgfplots}
\pgfplotsset{compat=1.17} 

\allowdisplaybreaks[4]

\def\Eavg{{\overline{E}}}
\def\Emin{{E_\text{min}}}
\def\Gapavg{{\overline{\Delta\epsilon}}}
\def\Gapmin{{\Delta\epsilon_\text{min}}}
\def\Gapmax{{\Delta\epsilon_\text{max}}}

\newcommand{\CVGAE}{\textsc{CVGAE}\xspace}
\newcommand{\GraphDG}{\textsc{GraphDG}\xspace}
\newcommand{\CGCF}{\textsc{CGCF}\xspace}
\newcommand{\ConfVAE}{\textsc{ConfVAE}\xspace}
\newcommand{\ConfGF}{\textsc{ConfGF}\xspace}

\newcommand{\GeoMol}{\textsc{GeoMol}\xspace}
\newcommand{\RDKit}{\textsc{RDKit}\xspace}
\newcommand{\method}{\textsc{GeoDiff}\xspace}
\newcommand{\Psifour}{\textsc{Psi4}\xspace}

\newcommand{\Jian}[1]{\textcolor{red}{[Jian: #1]}}

\newcommand{\kl}[2]{\mathrm{KL}\left(#1\|#2\right)}

\newtheorem{proposition}{Proposition}
\newtheorem{property}{Property}

\title{\method: a Geometric Diffusion Model for \\Molecular Conformation Generation}

\author{Minkai Xu$^{1,2}$, Lantao Yu$^{3}$, Yang Song$^{3}$, Chence Shi$^{1,2}$, Stefano Ermon$^{3*}$, Jian Tang$^{1,4,5*}$ \\
$^{1}$Mila - Qu\'ebec AI Institute, Canada $^{2}$Universit\'e de Montr\'eal, Canada\\
$^{3}$Stanford University, USA $^{4}$HEC Montr\'eal, Canada $^{5}$CIFAR AI Research Chair \\
\texttt{\{minkai.xu,chence.shi\}@umontreal.ca}\\
\texttt{\{lantaoyu,yangsong,ermon\}@cs.stanford.edu}\\
\texttt{jian.tang@hec.ca}
}

\iclrfinalcopy 
\begin{document}

\maketitle

\begin{abstract}

Predicting molecular conformations from molecular graphs is a fundamental problem in cheminformatics and drug discovery. Recently, significant progress has been achieved with machine learning approaches, especially with deep generative models. Inspired by the diffusion process in classical non-equilibrium thermodynamics where heated particles will diffuse from original states to a noise distribution, in this paper, we propose a novel generative model named \method for molecular conformation prediction. \method treats each atom as a particle and learns to directly reverse the diffusion process (\textit{i.e.}, transforming from a noise distribution to stable conformations) as a Markov chain. Modeling such a generation process is however very challenging as the likelihood of conformations should be roto-translational invariant. We theoretically show that Markov chains evolving with \textit{equivariant} Markov kernels can induce an \textit{invariant} distribution by design, and further propose building blocks for the Markov kernels to preserve the desirable equivariance property. The whole framework can be efficiently trained in an end-to-end fashion by optimizing a weighted variational lower bound to the (conditional) likelihood. Experiments on multiple benchmarks show that \method is superior or comparable to existing state-of-the-art approaches, especially on large molecules.\footnote{Code is available at \url{https://github.com/MinkaiXu/GeoDiff}.}



\end{abstract}
\section{Introduction}

Graph representation learning has achieved huge success for molecule modeling in various tasks ranging from property prediction~\citep{gilmer2017neural,duvenaud2015convolutional} to molecule generation~\citep{jin2018junction,shi2020graphaf}, where typically a molecule is represented as an atom-bond graph. 
Despite its effectiveness in various applications, a more intrinsic and informative representation for molecules is the 3D \textit{geometry}, also known as \textit{conformation}, where atoms are represented as their Cartesian coordinates. The 3D structures determine the biological and physical properties of molecules and hence play a key role in many applications such as computational drug and material design~\citep{Thomas2018TensorFN,gebauer2021inverse,jing2021gvp,batzner2021se}. Unfortunately, how to predict stable molecular conformation remains a challenging problem. Traditional methods based on molecular dynamics (MD) or Markov chain Monte Carlo (MCMC) are very computationally expensive, especially for large molecules~\citep{hawkins2017conformation}.

Recently, significant progress has been made with machine learning approaches, especially with deep generative models. 
For example, \cite{simm2020GraphDG,xu2021end} studied predicting atomic distances with variational autoencoders (VAEs)~\citep{kingma2013auto} and flow-based models~\citep{dinh2016density} respectively. \cite{shi*2021confgf} proposed to use denoising score matching~\citep{song2019generative,song2020scoretech} to estimate the gradient fields over atomic distances, through which the gradient fields over atomic coordinates can be calculated. \cite{ganea2021geomol} studied generating conformations by predicting both bond lengths and angles. As molecular conformations are roto-translational invariant, these approaches circumvent directly modeling atomic coordinates by leveraging intermediate geometric variables such as atomic distances, bond and torsion angles, which are roto-translational invariant. As a result, they are able to achieve very compelling performance. However, as all these approaches seek to indirectly model the intermediate geometric variables, they have inherent limitations in either training or inference process (see Sec.~\ref{sec:related} for a detailed description). Therefore, an ideal solution would still be directly modeling the atomic coordinates and at the same time taking the roto-translational invariance property into account.

In this paper, we propose such a solution called \method, a principled probabilistic framework based on denoising diffusion models~\citep{sohl2015deep}. 
Our approach is inspired by the \textit{diffusion process} in nonequilibrium thermodynamics~\citep{de2013non}.
We view atoms as particles in a thermodynamic system, 
which gradually diffuse from the original states to a noisy distribution in contact with a heat bath. 
At each time step, stochastic noises are added to the atomic positions. 
Our high-level idea is learning to reverse the diffusion process, which recovers the target geometric distribution from the noisy distribution.
In particular, inspired by recent progress of denoising diffusion models on image generation~\citep{ho2020denoising,song2020denoising}, we view the noisy geometries at different timesteps as latent variables, and formulate both the forward diffusion and reverse denoising process as Markov chains. 
Our goal is to learn the transition kernels such that the reverse process can recover realistic conformations from the chaotic positions sampled from a noise distribution.
However, extending existing methods to geometric generation is highly non-trivial:
a direct application of diffusion models on the conformation generation task lead to poor generation quality. As mentioned above, molecular conformations are roto-translational invariant, \textit{i.e.}, the estimated (conditional) likelihood should be unaffected by translational and rotational transformations~\citep{kohler20eqflow}. 
To this end, we first theoretically show that a Markov process starting from an roto-translational \textit{invariant} prior distribution and evolving with roto-translational \textit{equivariant} Markov kernels can induce an roto-translational \textit{invariant} density function. We further provide practical parameterization to define a roto-translational \textit{invariant} prior distribution and a Markov kernel imposing the equivariance constraints.
In addition, we derive a weighted variational lower bound of the conditional likelihood of molecular conformations, which also enjoys the roto-translational invariance and can be efficiently optimized. 

A unique strength of \method is that it directly acts on the atomic coordinates and entirely bypasses the usage of intermediate elements for both training and inference. 
This general formulation enjoys several crucial advantages. 
First, the model can be naturally trained end-to-end without involving any sophisticated techniques like bilevel programming~\citep{xu2021end}, which benefits from small optimization variances. 
Besides, instead of solving geometries from bond lengths or angles, the one-stage sampling fashion avoids accumulating any intermediate error, and therefore leads to more accurate predicted structures.
Moreover, \method enjoys a high model capacity to approximate the complex distribution of conformations.
Thus, the model can better estimate the highly multi-modal distribution and generate structures with high quality and diversity.

We conduct comprehensive experiments on multiple benchmarks, including conformation generation and property prediction tasks. 
Numerical results show that \method consistently outperforms existing state-of-the-art machine learning approaches, and by a large margin on the more challenging large molecules. 
The significantly superior performance demonstrate the high capacity to model the complex distribution of molecular conformations and generate both diverse and accurate molecules.

\section{Related Work}
\label{sec:related}

Recently, various deep generative models have been proposed for conformation generation. 
Among them, \CVGAE~\citep{mansimov19molecular} first proposed a VAE model to directly generate 3D atomic coordinates, which fails to preserve the roto-translation equivariance property
of conformations and suffers from poor performance.
To address this problem, the majority of subsequent models are based on intermediate geometric elements such as atomic distances and torsion angles.
A favorable property of these elements is the roto-translational invariance, (\textit{e.g.} atomic distances does not change when rotating the molecule), which has been shown to be an important inductive bias for molecular geometry modeling~\citep{kohler20eqflow}.
However, such a decomposition suffers from several drawbacks for either training or sampling. 
For example, \GraphDG~\citep{simm2020GraphDG} and \CGCF~\citep{xu2021cgcf} proposed to predict the interatomic distance matrix by VAE and Flow respectively, and then solve the geometry through the Distance Geometry (DG) technique~\citep{liberti2014euclidean}, which searches reasonable coordinates that matches with the predicted distances. 
\ConfVAE further improves this pipeline by designing an end-to-end framework via bilevel optimization~\citep{xu2021end}.
However, all these approaches suffer from the accumulated error problem, meaning that the noise in the predicted distances will misguide the coordinate searching process and lead to inaccurate or even erroneous structures. 
To overcome this problem, \ConfGF~\citep{shi*2021confgf,luo2021predicting} proposed to learn the gradient of the log-likelihood \textit{w.r.t} coordinates.
However, in practice the model is still aided by intermediate geometric elements, in that it first estimates the gradient \textit{w.r.t} interatomic distances via denoising score matching (DSM)~\citep{song2019generative,song2020scoretech},
and then derives the gradient of coordinates using the chain rule.
The problem is,
by learning the distance gradient via DSM, the model is fed with perturbed distance matrices, which may violate the triangular inequality or even contain negative values. 
As a consequence, the model is actually learned over invalid distance matrices but tested with valid ones calculated from coordinates, making it suffer from serious out-of-distribution~\citep{hendrycks2016baseline} problem.
Most recently, another concurrent work \citep{ganea2021geomol} proposed a highly \textit{systematic} (rule-based) pipeline named \GeoMol, which learns to predict a minimal set of geometric quantities (\textit{i.e.} length and angles) and then reconstruct the local and global structures of the conformation in a sophisticated procedure.
Besides, there has also been efforts to use reinforcement learning for conformation search~\cite{Gogineni2020TorsionNet}. Nevertheless, this method relies on rigid rotor approximation and can only model the torsion angles, and thus fundamentally differs from other approaches. 

\section{Preliminaries}
\subsection{Notations and Problem Definition}
\textbf{Notations.}
In this paper each molecule with $n$ atoms is represented as an undirected graph $\gG = \langle \gV, \gE \rangle$, where $\gV = \{ v_i \}_{i=1}^n$ is the set of vertices representing atoms and $\gE = \{ e_{ij} \mid (i, j) \subseteq |\gV| \times |\gV| \}$ is the set of edges representing inter-atomic bonds. Each node $v_i \in \gV$ describes the atomic attributes, \textit{e.g.}, the element type. 
Each edge $e_{ij} \in \gE$ describes the corresponding connection between $v_i$ and $v_j$, and is labeled with its chemical type. In addition, we also assign the unconnected edges with a \textit{virtual} type. For the geometry, each atom in $\gV$ is embedded by a coordinate vector $\vc \in \sR^3$ into the 3-dimensional space, and the full set of positions (\textit{i.e.}, the conformation) can be represented as a matrix $\gC = [\vc_1, \vc_2, \cdots , \vc_{n}] \in \sR^{n \times 3}$.

\textbf{Problem Definition.}
The task of \textit{molecular conformation generation} is a conditional generative problem, where we are interested in generating stable conformations for a provided graph $\gG$. Given multiple graphs $\gG$, and for each $\gG$ given its conformations $\gC$ as \textit{i.i.d} samples from an underlying Boltzmann distribution~\citep{noe2019boltzmann}, our goal is learning a generative model $p_\theta(\gC|\gG)$, which is easy to draw samples from, to approximate the Boltzmann function.

\subsection{Equivariance}
\label{subsec:equivariance}

\textit{Equivariance} is ubiquitous in machine learning for atomic systems, \textit{e.g.}, the vectors of atomic dipoles or forces should rotate accordingly \textit{w.r.t.} the conformation coordinates~\citep{Thomas2018TensorFN, Weiler20183DSC, fuchs2020se3, Miller2020RelevanceOR, simm2021symmetryaware,batzner2021se}. 
It has shown effectiveness to integrate such inductive bias into model parameterization for modeling 3D geometry, which is critical for the generalization capacity~\citep{kohler20eqflow,satorras2021enflow}.
Formally, a function $\gF: \gX \rightarrow \gY$ 
is equivariant \textit{w.r.t} a group $G$ if:
\begin{equation}
\label{eq:equivariance}
    \mathcal{F} \circ T_g (x) = S_g \circ \mathcal{F} (x),
\end{equation}
where $T_g$ and $S_g$ are transformations for an element $g \in G$, acting on the vector spaces $\gX$ and $\gY$, respectively. 
In this work, 
we consider the SE(3) group, \textit{i.e.}, 
the group of rotation, translation in 3D space.
This requires the estimated likelihood unaffected with translational and rotational transformations,
and we will elaborate on how our method satisfy this property in Sec.~\ref{sec:method}.

\section{GeoDiff Method}
\label{sec:method}

In this section, we elaborate on the proposed equivariant diffusion framework. We first present a high level description of our 3D diffusion formulation in Sec.~\ref{subsec:overview}, based on recent progress of denoising diffusion models~\citep{sohl2015deep,ho2020denoising}.
Then we emphasize several non-trivial challenges of building diffusion models for geometry generation scenario, and show how we technically tackle these issues.
Specifically, in Sec.~\ref{subsec:equivariant-decoder}, we present how we parameterize $p_\theta(\gC|\gG)$ so that the conditional likelihood is roto-translational invariant, and in Sec.~\ref{subsec:equivariant-loss}, we introduce our surgery of the training objective to make the optimization also invariant of translation and rotation. Finally, we briefly show how to draw samples from our model in Sec.~\ref{subsec:sampling}. 

\begin{figure}[t]
	\centering
    \includegraphics[width=.8\linewidth]{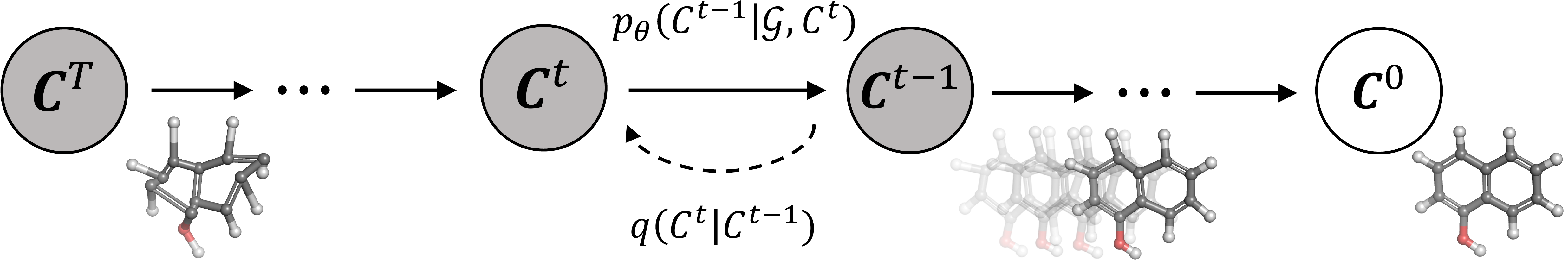}
    \caption{Illustration of the diffusion and reverse process of \method. 
    For diffusion process, noise from fixed posterior distributions $q(\gC^{t}|\gC^{t-1})$ is gradually added until the conformation is destroyed.
    Symmetrically, for generative process, an initial state $\gC^{T}$ is sampled from standard Gaussian distribution, and the conformation is progressively refined via the Markov kernels $p_{\theta}(\gC^{t-1}|\gG,\gC^t)$.
    }
    \label{fig:framework}
\end{figure}

\subsection{Formulation}
\label{subsec:overview}


Let $\gC^0$
denotes the ground truth conformations and let $\gC^t$ for $t=1,\cdots,T$ be a sequence of latent variables with the same dimension, where $t$ is the index for diffusion steps. Then a diffusion probabilistic model~\citep{sohl2015deep} can be described as a latent variable model with two processes: the forward \textit{diffusion} process, and the reverse \textit{generative} process.
Intuitively, the \textit{diffusion process} progressively injects small noises to the data $\gC^{0}$,
while the \textit{generative process} learns to revert the diffusion process by gradually eliminating the noise to recover the ground truth.
We provide a high-level schematic of the processes in Fig.~\ref{fig:framework}.

\textbf{Diffusion process.}
Following the physical insight, we model the particles $\gC$ as an evolving thermodynamic system. With time going by, the equilibrium conformation $\gC^0$ will gradually diffuse to the next chaotic states $\gC^t$, and finally converge into a white noise distribution after $T$ iterations. 
Different from typical latent variable models, in diffusion model this \textit{forward process} is defined as a fixed (rather than trainable) posterior distribution $q(\gC^{1:T}|\gC^0)$.
Specifically, we define it as a Markov chain according to a fixed variance schedule $\beta_1,\dots,\beta_T$:
\begin{equation}
\label{eq:diffusion}
    q(\gC^{1:T}|\gC^0) = \prod_{t=1}^T q(\gC^t|\gC^{t-1}), \quad q(\gC^t|\gC^{t-1}) = \gN(\gC^t;\sqrt{1-\beta_t}\gC^{t-1},\beta_t I).
\end{equation}
Note that, 
in this work we do not impose specific (invariance) requirement upon the diffusion process, as long as it can efficiently draw noisy samples for training the generative process $p_\theta(\gC^{0})$. 

Let $\alpha_t = 1 - \beta_t$ and $\bar{\alpha}_t = \prod_{s=1}^t\alpha_s$, a special property of the forward process is that $q(\gC^t|\gC^0)$ of arbitrary timestep $t$ can be calculated in closed form $q(\gC^t|\gC^0)=\gN(\gC^t;\sqrt{\bar{\alpha}_t}\gC^{0},(1-\bar{\alpha}_t) I)$\footnote{Detailed derivations are provided in the Appendix~\ref{app:sec:proof}.}. This indicates with sufficiently large $T$, the whole forward process will convert $\gC^0$ to whitened isotropic Gaussian, and thus it is natural to set $p(\gC^T)$ as a standard Gaussian distribution.

    


\textbf{Reverse Process.} Our goal is learning to recover conformations $\gC^0$ from the white noise $\gC^T$, given specified molecular graphs $\gG$.
We consider this generative procedure as a reverse dynamics of the above diffusion process, starting from the noisy particles $\gC^T \sim p(\gC^T)$. 
We formulate this reverse dynamics as a conditional Markov chain with learnable transitions:
\begin{equation}
\label{eq:reverse-markov}
    p_{\theta}(\gC^{0:{T-1}}|\gG, \gC^T)=\prod_{t=1}^T p_{\theta}(\gC^{t-1}|\gG,\gC^t), \quad p_{\theta}(\gC^{t-1}|\gG,\gC^t) = \gN(\gC^{t-1};\mu_{\theta}(\gG, \gC^t,t), \sigma^2_t I).
\end{equation}
Herein $\mu_\theta$ are parameterized neural networks to estimate the means, and $\sigma_t$ can be any user-defined variance.
The initial distribution $p(\gC^T)$ is set as a standard Gaussian. 
Given a graph $\gG$, its 3D structure is generated by first drawing chaotic particles $\gC^T$ from $p(\gC^T)$, and then iteratively refined through the reverse Markov kernels $p_{\theta}(\gC^{t-1}|\gG,\gC^t)$.

Having formulated the \textit{reverse} dynamics, the marginal likelihood can be calculated by $p_{\theta}(\gC^{0}|\gG)=\int p(\gC^T) p_{\theta}(\gC^{0: T-1}|\gG,\gC^T) \mathrm{d} \boldsymbol{\gC}^{1: T}$.
Herein a non-trivial problem is that the likelihood should be invariant \textit{w.r.t} translation and rotation, which has proved to be a critical inductive bias for 3D object generation~\citep{kohler20eqflow,satorras2021enflow}.
In the following subsections, we will elaborate on how we 
parameterize the Markov kernels $p_\theta(\gC^{t-1}|\gG,\gC^t)$ to achieve this desired property, and also how to maximize this likelihood by taking the invariance into account. 


\subsection{Equivariant Reverse Generative Process}
\label{subsec:equivariant-decoder}




Instead of directly leveraging existing methods, we consider building the density $p_\theta(\gC^0)$ that is invariant to rotation and translation transformations.
Intuitively, this requires the likelihood to be unaffected by translations and rotations. 
Formally, let $T_g$ be some roto-translational transformations of a group element $g \in$ SE(3), then we have the following statement:
\begin{proposition}
\label{prop:markov-equivariance}
    Let $p(x_T)$ be an SE(3)-invariant density function, \textit{i.e.}, $p(x_T)=p(T_g (x_T))$. If Markov transitions $p(x_{t-1}|x_t)$ are SE(3)-equivariant, \textit{i.e.}, $p(x_{t-1}|x_t) = p(T_g (x_{t-1})|T_g (x_t))$, then we have that the density $p_{\theta}(x_{0})=\int p(x_T) p_{\theta}(x_{0: T-1}|x_T) \mathrm{d} \boldsymbol{x}_{1: T}$ is also SE(3)-invariant.
\end{proposition}
This proposition indicates that the dynamics starting from an invariant standard density along an equivariant Gaussian Markov kernel can result in an invariant density. 
Now we provide a practical implementation of \method based on the recent \textit{denoising diffusion} framework~\citep{ho2020denoising}.

\textbf{Invariant Initial Density $p(\gC^T)$.} We first introduce the invariant distribution $p(\gC^T)$, which will also be employed in the equivariant Markov chain. We borrow the idea from \cite{kohler20eqflow} to consider systems with zero center of mass (CoM), termed CoM-free systems. We define $p(\gC^T)$ as a ``CoM-free standard density'' $\hat{\rho}(\gC)$, built upon an isotropic normal density $\rho(\gC)$: for evaluating the likelihood $\hat{\rho}(\gC)$ we can firstly translate $\gC$ to zero CoM and then calculate $\rho(\gC)$, and for sampling from $\hat{\rho}(\gC)$ we can first sample from $\rho(\gC)$ and then move the CoM to zero. 

We provide a formal theoretical analysis of $\hat{\rho}(\gC)$ in Appendix~\ref{app:sec:proof}. Intuitively, the isotropic Gaussian is manifestly invariant to rotations around the zero CoM. And by considering CoM-free system, moving the particles to zero CoM can always ensure the translational invariance. Consequently, $\hat{\rho}(\gC)$ is constructed as a roto-transitional invariant density.

\textbf{Equivariant Markov Kernels $p(\gC^{t-1} | \gG, \gC^t)$.}
Similar to the prior density, we also consider equipping all intermediate structures $\gC^t$ as CoM-free systems.
Specifically, given mean $\mu_\theta(\gG, \gC^t, t)$ and variance $\sigma_t$, the likelihood of $\gC^{t-1}$ will be calculated by $\hat{\rho}(\frac{\gC^{t-1} - \mu_\theta(\gG, \gC^t, t)}{\sigma_t})$. The CoM-free Gaussian ensures the translation invariance in the Markov kernels. Consequently, to achieve the equivariant property defined in Proposition~\ref{prop:markov-equivariance}, we focus on the rotation equivariance. 

Then in general, the key requirement is to ensure the means $\mu_\theta (\gG, \gC^t, t)$ to be roto-translation equivariant \textit{w.r.t} $\gC^t$. Following \cite{ho2020denoising}, we consider the following parameterization of $\mu_\theta$:
\begin{equation}
\label{eq:parameterization}
\mu_{\theta}(\gC^t, t) = \frac{1}{\sqrt{\alpha_t}}\left(\gC^t-\frac{\beta_t}{\sqrt{1-\bar{\alpha}_t}}\epsilon_{\theta}(\gG, \gC^t, t)\right), 
\end{equation}
where $\epsilon_{\theta}$ are neural networks with trainable parameters $\theta$.
Intuitively, the model $\epsilon_\theta$ learns to predict the noise necessary to decorrupt the conformations. This is analogous to the physical force fields~\citep{schutt2017schnet,zhang2018deepmd,hu2021forcenet,shuaibi2021rotation}, which also gradually push particles towards convergence around the  equilibrium states. 

Now the problem is transformed to constructing $\epsilon_\theta$ to be roto-translational equivariant.
We draw inspirations from recent equivariant networks~\citep{Thomas2018TensorFN,satorras2021en} to design an equivariant convolutional layer, named graph field network (GFN). In the $l$-th layer, GFN takes node embeddings $\rvh^l \in \mathbb{R}^{n \times b}$ ($b$ denotes the feature dimension) and corresponding coordinate embeddings $\rvx^l \in \mathbb{R}^{n \times 3}$ as inputs, and outputs $\rvh^{l+1}$ and $\rvx^{l+1}$ as follows:
\begin{align}
    & \rvm_{ij} =\Phi_{m}\left(\rvh_{i}^{l}, \rvh_{j}^{l},\|\rvx_{i}^{l}-\rvx_{j}^{l}\|^{2}, e_{ij}; \theta_m \right) \label{eq:nfl-message} \\
    & \rvh_{i}^{l+1} =\Phi_{h}\Big(\rvh_{i}^l, \sum_{j \in \mathcal{N}(i)} \rvm_{ij}; \theta_h \Big) \label{eq:nfl-node} \\
    & \rvx_{i}^{l+1} = \sum_{j \in \mathcal{N}(i)}\frac{1}{d_{ij}}\left(\rvc_{i}-\rvc_{j}\right) \Phi_{x}\left(\rvm_{ij} ; \theta_x \right) \label{eq:nfl-tensor}
\end{align}
where $\Phi$ are feed-forward networks and $d_{ij}$ denotes interatomic distances.
$\gN(i)$ denotes the neighborhood of $i^{th}$ node, including both connected atoms and other ones within a radius threshold $\tau$, which enables the model to explicitly capture long-range interactions and support molecular graphs with disconnected components.
Initial embeddings $\rvh^0$ are combinations of atom and timestep embeddings, and $\rvx^0$ are atomic coordinates. The main difference between proposed GFN and other GNNs lies in \eqref{eq:nfl-tensor}, where $\rvx$ is updated as a combination of radial directions weighted by $\Phi_x:\mathbb{R}^b \rightarrow \mathbb{R}$. Such vector field $\rvx^L$ enjoys the roto-translation equivariance property. Formally, we have:
\begin{proposition}
\label{prop:gfn-equivariant}
    Parameterizing $\epsilon_\theta(\gG,\gC,t)$ as a composition of $L$ GFN layers, and take the $\rvx^L$ after $L$ updates as the output. Then the noise vector field $\epsilon_\theta$ is SE(3) equivariant \textit{w.r.t} the 3D system $\gC$.
\end{proposition}
Intuitively, given $\rvh^l$ already invariant and $\rvx^l$ equivariant, the message embedding $\rvm$ will also be invariant since it only depends on  invariant features. Since $\rvx$ is updated with the relative differences $\rvc_{i}-\rvc_{j}$ weighted by invariant features, it will be translation-invariant and rotation-equivariant. Then inductively, composing $\epsilon_\theta$ with $L$ GFN layers enables equivariance with $\gC^t$. We provide the formal proof of equivariance properties in Appendix~\ref{app:sec:proof}.


\subsection{Improved Training Objective}
\label{subsec:equivariant-loss}

%
Having formulated the generative process and the model parameterization, now we consider the practical training objective for the reverse dynamics. 
Since directly optimizing the exact log-likelihood is intractable, we instead maximize the usual variational lower bound (ELBO)\footnote{The detailed derivations and full proofs are provided in Appendix~\ref{app:sec:proof}. \label{footnote:diffusion-object}}:
\begin{align}
\label{eq:elbo}
    \mathbb{E} \left[\log p_{\theta}(\gC^0|\gG) \right]
    &=  \mathbb{E} \Big[ \log\mathbb{E}_{q(\gC^{1:T}|\gC^0)} \frac{p_{\theta}(\gC^{0:T}|\gG)}{q(\gC^{1:T}|\gC^0)} \Big] \nonumber \\
    &\geq - \mathbb{E}_{q} \Big[ 
    \sum_{t = 1}^T
    D_{\text{KL}}(q(\gC^{t-1}|\gC^t,\gC^0)\|p_\theta(\gC^{t-1}|\gC^t,\gG))
    \Big] 
    := - \gL_\mathrm{ELBO}
\end{align}
where $q(\gC^{t-1}|\gC^t,\gC^0)$ is analytically tractable as $\gN(\frac{\sqrt{\bar{\alpha}_{t-1}} \beta_{t}}{1-\bar{\alpha}_{t}} \gC^{0}+\frac{\sqrt{\alpha_{t}}\left(1-\bar{\alpha}_{t-1}\right)}{1-\bar{\alpha}_{t}} \gC^{t}, \frac{1-\bar{\alpha}_{t-1}}{1-\bar{\alpha}_{t}} \beta_{t})$\footref{footnote:diffusion-object}.
Most recently, \citet{ho2020denoising} showed that under the parameterization in \eqref{eq:parameterization}, the ELBO of the diffusion model can be further simplified by calculating the KL divergences between Gaussians as weighted $\gL_2$ distances between the means $\epsilon_\theta$ and $\epsilon$\footref{footnote:diffusion-object}. Formally, we have:
\begin{proposition}\citep{ho2020denoising}
\label{prop:elbo}
Under the parameterization in \eqref{eq:parameterization}, we have:
\begin{equation}
\label{eq:train-obj}
\gL_\mathrm{ELBO} = \sum_{t=1}^T\gamma_t\mathbb{E}_{\{\gC^0,\gG\} \sim q(\gC^0,\gG),\epsilon\sim\gN(0,I)}\Big[ \|\epsilon-\epsilon_{\theta}(\gG,\gC^t,t)\|_2^2 \Big]
\end{equation}
where $\gC^t=\sqrt{\bar{\alpha}_t}\gC^0+\sqrt{1-\bar{\alpha}_t}\epsilon$. The weights $\gamma_t=\frac{\beta_t}{2\alpha_t(1-\bar{\alpha}_{t-1})}$ ~for~ $t>1$, and $\gamma_1=\frac{1}{2\alpha_1}$. 
\end{proposition}
%
The intuition of this objective is to independently sample chaotic conformations of different timesteps from $q(\gC^{t-1}|\gC^t,\gC^0)$, and use $\epsilon_\theta$ to model the noise vector $\epsilon$.
To yield a better empirical performance, \cite{ho2020denoising} suggests to set all weights $\gamma_t$ as $1$, which is in line with the the objectives of recent noise conditional score networks~\citep{song2019generative,song2020scoretech}.

As $\epsilon_\theta$ is designed to be equivariant, it is natural to require its supervision signal $\epsilon$ to be equivariant with $\gC^t$. 
Note that once this is achieved, the ELBO will also become invariant.
However, the $\epsilon$ in the forward diffusion process is not imposed with such equivariance, violating the above properties.
Here we propose two approaches to obtain the modified noise vector $\hat{\epsilon}$, which, after replacing $\epsilon$ in the $\gL_2$ distance calculation in \eqref{eq:train-obj}, achieves the desired equivariance:

\textbf{Alignment approach}. Considering the fact that $\epsilon$ can be calculated by $\frac{\gC^t-\sqrt{\bar{\alpha}_t}\gC^0}{\sqrt{1-\bar{\alpha}_t}}$, we can first rotate and translate $\gC^0$ to $\hat{\gC}^0$ by aligning \textit{w.r.t} $\gC^t$, and then compute $\hat{\epsilon}$ as $\frac{\gC^t-\sqrt{\bar{\alpha}_t}\hat{\gC}^0}{\sqrt{1-\bar{\alpha}_t}}$. Since the aligned conformation $\hat{\gC}^0$ is equivariant with $\gC^t$, the processed $\hat{\epsilon}$ will also enjoy the equivariance.
Specifically, 
the alignment is implemented by first translating $\gC^0$ to the same CoM of $\gC^t$ and then solve the optimal rotation matrix by Kabsch alignment algorithm~\citep{kabsch1976solution}. 

\textbf{Chain-rule approach}. Another meaningful observation is that by reparameterizing the Gaussian distribution $q(\gC^t|\gC^0)$ as $\gC^t=\sqrt{\bar{\alpha}_t}\gC^0+\sqrt{1-\bar{\alpha}_t}\epsilon$, 
$\epsilon$ can be viewed as a weighted score function  ${\sqrt{1-\bar{\alpha}_t}}\nabla_{\gC^{t}}\, q(\gC^t|\gC^0)$. \cite{shi*2021confgf} recently shows that generally this score function $\nabla_{\gC^t}\, q(\gC^t|\cdot)$ can be designed to be equivariant by decomposing it into $\partial_{\gC^t} \rvd^t \, \nabla_{\rvd^t} q(\gC^t|\cdot)$ with the chain rule, where $\rvd^t$ can be any invariant features of the structures $\gC^t$ such as the inter-atomic distances. We refer readers to \cite{shi*2021confgf} for more details. The insight is that as gradient of invariant variables $w.r.t$ equivariant variables, the partial derivative $\partial_{\gC^t} \rvd^t$ will always be equivalent with $\gC^t$. In this work, under the common assumption that $\rvd$ also follows a Gaussian distribution~\citep{kingma2013auto}, our practical implementation is to first approximately calculate $\nabla_{\rvd^t} q(\gC^t|\gC^0)$ as $\frac{\rvd^t - \sqrt{\bar{\alpha}_t} \rvd^0}{1-\bar{\alpha}_t}$,
and then compute the modified noise vector $\hat \epsilon$ as ${\sqrt{1-\bar{\alpha}_t}}\, \partial_{\gC^t} \rvd^t (\frac{\rvd^t - \sqrt{\bar{\alpha}_t} \rvd^0}{1-\bar{\alpha}_t})= \frac{\partial_{\gC^t} \rvd^t \cdot (\rvd^t - \sqrt{\bar{\alpha}_t} \rvd^0)}{\sqrt{1-\bar{\alpha}_t}}$.

\subsection{Sampling}
\label{subsec:sampling}

\begin{wrapfigure}[12]{R}{0.65\textwidth}
\vspace{-20pt}
\begin{minipage}{0.65\textwidth}
\begin{algorithm}[H]
    \caption{Sampling Algorithm of \method.}
    \label{alg:sampling}
    \textbf{Input}: the molecular graph $\gG$, the learned reverse model $\epsilon_\theta$.\\
    \textbf{Output}: the molecular conformation $\gC$.
    
    \begin{algorithmic}[1]
        \STATE Sample $\gC^{T} \sim p(\gC^T)= \gN(0,I)$
        \FOR{$s=T,T-1,\cdots,1$}
            \STATE Shift $\gC^{s}$ to zero CoM
            \STATE Compute $\mu_{\theta}(\gC^s,\gG,s)$ from $\epsilon_{\theta}(\gC^s,\gG,s)$ using \eqref{eq:parameterization}
            \STATE Sample $\gC^{s-1} \sim \gN(\gC^{s-1}; \mu_{\theta}(\gC^s,\gG,s), \sigma_t^2 I)$
        \ENDFOR
        \RETURN{$\gC^0$ as $\gC$}
    \end{algorithmic}
\end{algorithm}
\end{minipage}
\end{wrapfigure}

With a learned reverse dynamics $\epsilon_\theta(\gG,\gC^t,t)$, the transition means $\mu_\theta(\gG,\gC^t,t)$ can be calculated by \eqref{eq:parameterization}.
Thus, given a graph $\gG$, its geometry $\gC^0$ is generated by first sampling chaotic particles $\gC^T \sim p(\gC^T)$, and then progressively sample $\gC^{t-1} \sim p_{\theta}(\gC^{t-1}|\gG,\gC^t)$ for $t=T,T-1,\cdots,1$. 
This process is Markovian, which gradually shifts the previous noisy positions towards equilibrium states. 
We provide the pseudo code of the whole sampling process in Algorithm~\ref{alg:sampling}.

\section{Experiment}

In this section, we empirically evaluate \method on the task of equilibrium conformation generation for both small and drug-like molecules. 
Following existing work~\citep{shi*2021confgf,ganea2021geomol}, we test the proposed method as well as the competitive baselines on two standard benchmarks: \textbf{Conformation Generation} (Sec.~\ref{subsec:exp:conf}) and \textbf{Property Prediction} (Sec.~\ref{subsec:exp:prop}). We first present the general experiment setups, and then describe task-specific evaluation protocols and discuss the results in each section. 
The implementation details are provided in Appendix~\ref{app:sec:train-details}.

\subsection{Experiment Setup}

\textbf{Datasets.} Following prior works~\citep{xu2021cgcf,xu2021end}, we also use the recent GEOM-QM9~\citep{ramakrishnan2014quantum} and GEOM-Drugs~\citep{axelrod2020geom} datasets. The former one contains small molecules while the latter one are medium-sized organic compounds. We borrow the data split produced by \cite{shi*2021confgf}. For both datasets, the training split consists of $40,000$ molecules with $5$ conformations for each, resulting in $200,000$ conformations in total. The valid split share the same size as training split. The test split contains $200$ distinct molecules, with $22,408$ conformations for QM9 and $14,324$ ones for Drugs.

\textbf{Baselines.}
We compare \method with $6$ recent or established state-of-the-art baselines. 
For the ML approaches, we test the following models with highest reported performance: \CVGAE~\citep{mansimov19molecular}, \GraphDG~\citep{simm2020GraphDG}, \CGCF~\citep{xu2021cgcf}, \ConfVAE~\citep{xu2021end} and \ConfGF~\citep{shi*2021confgf}. 
We also test the classic \RDKit~\citep{sereina2015rdkit} method, which is arguably the most popular open-source software for conformation generation.
We refer readers to Sec.~\ref{sec:related} for a detailed discussion of these models.

\subsection{Conformation Generation}
\label{subsec:exp:conf}

\textbf{Evaluation metrics.}
The task aims to measure both quality and diversity of generated conformations by different models. 
We follow \cite{ganea2021geomol} to evaluate $4$ metrics built upon root-mean-square deviation (RMSD), which is defined as the normalized Frobenius norm of two atomic coordinates matrices, after alignment by Kabsch algorithm~\citep{kabsch1976solution}. Formally, let $S_g$ and $S_r$ denote the sets of generated and reference conformers respectively, then the \textbf{Cov}erage and \textbf{Mat}ching metrics~\citep{xu2021cgcf} following the conventional \textit{Recall} measurement can be defined as:
\begin{align}
    \operatorname{COV-R}(S_g, S_r) &= \frac{1}{\vert S_r \vert} \Big\vert
    \Big\{\gC\in S_r \vert \operatorname{RMSD}(\gC, \hat{\gC}) \le \delta,  \hat{\gC} \in S_g\Big\}
    \Big\vert, \\
    \operatorname{MAT-R}(S_g, S_r) &= \frac{1}{\vert S_r \vert}
    \sum\limits_{\gC \in S_r}
    \min\limits_{\hat{\gC} \in S_g} \operatorname{RMSD}(\gC, \hat{\gC}),
\end{align}
where $\delta$ is a pre-defined threshold. The other two metrics COV-P and MAT-P inspired by \textit{Precision} can be defined similarly but with the generated and reference sets exchanged.
In practice, $S_g$ is set as twice of the size of $S_r$ for each molecule.
Intuitively, the COV scores measure the percentage of structures in one set covered by another set, where covering means the RMSD between two conformations is within a certain threshold $\delta$. 
By contrast, the MAT scores measure the average RMSD of conformers in one set with its closest neighbor in another set. 
In general, higher COV rates or lower MAT score suggest that more realistic conformations are generated.
Besides, the \textit{Precision} metrics depend more on the quality, while the \textit{Recall} metrics concentrate more on the diversity. Either metrics can be more appealing considering the specific scenario. Following previous works~\citep{xu2021cgcf,ganea2021geomol}, $\delta$ is set as $0.5 \si{\angstrom}$ and $1.25 \si{\angstrom}$ for QM9 and Drugs datasets respectively.

\begin{table}[!t]
    \vspace{-10pt}
    \centering
    \caption{Results on the \textbf{GEOM-Drugs} dataset,  without FF optimization.}
    \label{tab:drugs}
    \resizebox{\textwidth}{!}{
    \begin{threeparttable}
    \begin{tabular}{l|cccc|cccc}
    \toprule[1.0pt]
    & \multicolumn{2}{c}{\shortstack[c]{COV-R (\%) $\uparrow$}} & \multicolumn{2}{c|}{\shortstack[c]{MAT-R (\si{\angstrom}) $\downarrow$ }}  & \multicolumn{2}{c}{\shortstack[c]{COV-P (\%) $\uparrow$}}  & \multicolumn{2}{c}{\shortstack[c]{MAT-P (\si{\angstrom}) $\downarrow$ }} \\
    Models & Mean & Median & Mean & Median & Mean & Median & Mean & Median \\
    \midrule[0.8pt]
    \CVGAE & 0.00 & 0.00 & 3.0702 & 2.9937 & - & - & - & - \\ 
    \GraphDG & 8.27 & 0.00 & 1.9722 & 1.9845 & 2.08 & 0.00 & 2.4340 & 2.4100 \\ 
    \CGCF & 53.96 & 57.06 & 1.2487 & 1.2247 & 21.68 & 13.72 & 1.8571 & 1.8066 \\
    \ConfVAE & 55.20 & 59.43 & 1.2380 & 1.1417 & 22.96 & 14.05 & 1.8287 & 1.8159 \\
    \GeoMol & 67.16 & 71.71 & 1.0875 & 1.0586 & - & - & - & - \\ 
    \ConfGF & 62.15 & 70.93 & 1.1629 & 1.1596 & 23.42 & 15.52 & 1.7219 & 1.6863 \\
    \midrule[0.3pt]
    \textbf{\method-A} & 88.36 & 96.09 & 0.8704 & 0.8628 & 60.14 & 61.25 & 1.1864 & 1.1391 \\
    \textbf{\method-C} &  \bf 89.13 & \bf 97.88 & \bf 0.8629 & \bf 0.8529 & \bf 61.47 & \bf 64.55 & \bf 1.1712 & \bf 1.1232 \\
    \bottomrule[1.0pt]
    \end{tabular}
    \begin{tablenotes}
      \small
      \item * The COV-R and MAT-R results of \CVGAE, \GraphDG, \CGCF, and \ConfGF are borrowed from \cite{shi*2021confgf}. The results of \GeoMol are borrowed from a most recent study \cite{zhu2022direct}. Other results are obtained by our own experiments. The results of all models for the GEOM-QM9 dataset (summarized in Tab.~\ref{tab:qm9}) are collected in the same way.
    \end{tablenotes}
    \end{threeparttable}
    }
\end{table}

\begin{figure}[t]
	\centering
    \vspace{-5pt}    
    \resizebox{\textwidth}{!}{
    \includegraphics[width=1.\linewidth]{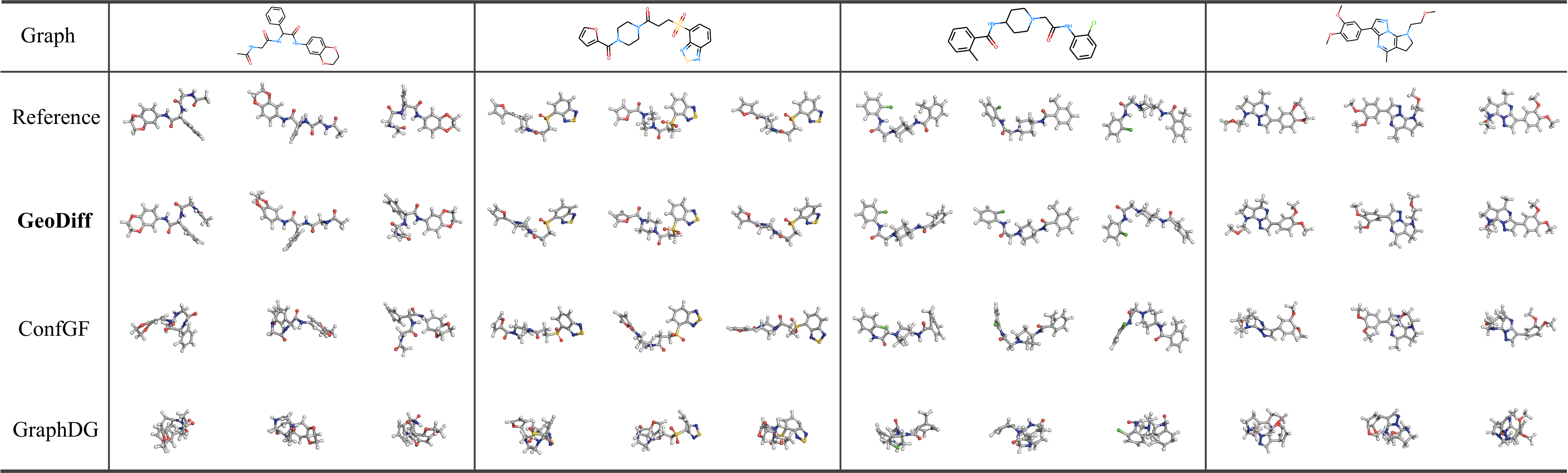}
    }
    \vspace{-20pt}
    \caption{Examples of generated structures from Drugs dataset. For every model, we show the conformation best-aligned with the ground truth. More examples are provided in Appendix~\ref{app:sec:sample}.}
    \label{fig:exp_drugs}
    \vspace{-10pt}
\end{figure}

\textbf{Results \& discussion.} 
The results are summarized in Tab.~\ref{tab:drugs} and Tab.~\ref{tab:qm9} (left in Appendix.~\ref{app:sec:exp}). 
As noted in Sec.~\ref{subsec:equivariant-loss}, \method can be trained with two types of modified ELBO, named \textit{alignment} and \textit{chain-rule} approaches. 
We denote models learned by these two objectives as \method-A and \method-C respectively. 
As shown in the tables, \method consistently outperform the state-of-the-art ML models on all datasets and metrics, especially by a significant margin for more challenging large molecules (Drugs dataset). 
The results demonstrate the superior capacity of \method to model the multi modal distribution, and generative both accurate and diverse conformations. 
We also notice that in general \method-C performs slightly better than \method-A, which suggests that \textit{chain-rule approach} leads to a better optimization procedure. We thus take \method-C as the representative in the following comparisons.
We visualize samples generated by different models in Fig.~\ref{fig:exp_drugs} to provide a qualitative comparison, where \method is shown to capture better both local and global structures.

On the more challenging Drugs dataset, we further test \RDKit.
As shown in Tab.~\ref{tab:rdkit}, our observation is in line with previous studies~\citep{shi*2021confgf} that the state-of-the-art ML models (shown in Tab.~\ref{tab:drugs}) perform better on COV-R and MAT-R. However, for the new \textit{Precision}-based metrics we found that ML models are still not comparable. This indicates that ML models tend to explore more possible representatives while \RDKit concentrates on a few most common ones, prioritizes quality over diversity. 
Previous works~\citep{mansimov19molecular,xu2021end} suggest that this is because \RDKit involves an additional empirical force field (FF)~\citep{halgren1996merck-extension} to optimize the structure, and we follow them to also combine \method with FF to yield a more fair comparison. Results in Tab.~\ref{tab:rdkit} demonstrate that \method+FF can keep the superior diversity (\textit{Recall} metrics) while also enjoy significantly improved accuracy ((\textit{Precision} metrics)). 

\begin{table}[t]
    \centering
    \small
    \caption{Results on the \textbf{GEOM-Drugs} dataset, with FF optimization.}
    \label{tab:rdkit}
    \resizebox{\linewidth}{!}{
    \begin{tabular}{l|cccc|cccc}
    \toprule[1.0pt]
     & \multicolumn{2}{c}{\shortstack[c]{COV-R (\%) $\uparrow$}} & \multicolumn{2}{c|}{\shortstack[c]{MAT-R (\si{\angstrom}) $\downarrow$ }}  & \multicolumn{2}{c}{\shortstack[c]{COV-P (\%) $\uparrow$}}  & \multicolumn{2}{c}{\shortstack[c]{MAT-P (\si{\angstrom}) $\downarrow$ }} \\
    Models & Mean & Median & Mean & Median & Mean & Median & Mean & Median \\
    \midrule[0.8pt]
    \RDKit & 60.91 & 65.70 & 1.2026 & 1.1252 & 72.22 & 88.72 & 1.0976 & 0.9539 \\ 
    \textbf{\method + FF} & \bf 92.27 & \bf 100.00 & \bf 0.7618 & \bf 0.7340 & \bf 84.51 & \bf 95.86 & \bf 0.9834 & \bf 0.9221 \\ 
    \bottomrule[1.0pt]
    \end{tabular}
    }
    \vspace{-5pt}
\end{table}

\subsection{Property Prediction}
\label{subsec:exp:prop}

\begin{wraptable}[9]{R}{0.55\textwidth}
\vspace{-20pt}
    \centering
    \caption{MAE of predicted ensemble properties in eV.}
    \label{tab:prop}
    \resizebox{\linewidth}{!}{
    \begin{tabular}{l | ccccc}
    \toprule
    Method & $ \overline{E} $ & $E_\text{min}$ & $ \overline{\Delta\epsilon} $ & $\Delta \epsilon_\text{min}$  & $\Delta \epsilon_\text{max}$ \\
    \midrule
    \RDKit & 0.9233 & 0.6585 & 0.3698 & 0.8021 & 0.2359 \\
    \GraphDG & 9.1027 & 0.8882 & 1.7973 & 4.1743 & 0.4776 \\
    \CGCF & 28.9661 & 2.8410 & 2.8356 & 10.6361 & 0.5954 \\
    \ConfVAE & 8.2080 & 0.6100 & 1.6080 & 3.9111 & 0.2429 \\
    \ConfGF & 2.7886 & 0.1765 & 0.4688 & 2.1843 & \bf 0.1433 \\
    \midrule
    \method & \bf 0.25974 & \bf 0.1551 & \bf 0.3091 & \bf 0.7033 & 0.1909 \\
    \bottomrule
    \end{tabular}
    }
\end{wraptable}

\textbf{Evaluation metrics.} 
This task estimates the molecular \textit{ensemble properties}~\citep{axelrod2020geom} over a set of generated conformations.
This can provide an direct assessment on the quality of generated samples.
In specific, we follow \cite{shi*2021confgf} to extract a split from GEOM-QM9 covering $30$ molecules, and generate $50$ samples for each. 
Then we use the chemical toolkit \Psifour~\citep{Smith2020Psi41O} to calculate each conformer's energy $E$ and HOMO-LUMO gap $\epsilon$, and compare the average energy $\Eavg$, lowest energy $\Emin$, average gap $\Gapavg$, minimum gap $\Gapmin$, and maximum gap $\Gapmax$ with the ground truth. 

\textbf{Results \& discussions.} The mean absolute errors (MAE) between calculated properties and the ground truth are reported in Tab.~\ref{tab:prop}.
\CVGAE is excluded due to the poor performance, which is also reported in \cite{simm2020GraphDG, shi*2021confgf}.
The properties are highly sensitive to geometric structure, and thus the superior performance demonstrate that \method can consistently predict more accurate conformations across different molecules.
\section{Conclusion}

We propose \method, a novel probabilistic model for generating molecular conformations.
\method marries denoising diffusion models with geometric representations, 
where we parameterize the reverse generative dynamics as a Markov chain, and novelly impose roto-translational invariance into the density with equivariant Markov kernels. 
We derive a tractable invariant objective from the variational lower bound to optimize the likelihood.
Comprehensive experiments over
multiple tasks demonstrate that \method is competitive with the existing state-of-the-art models. 
Future work includes further improving or accelerating the model with other recent progress of diffusion models, and extending our method to other challenging structures such as proteins.

\newpage
\section*{Acknowledgement}

Minkai thanks Huiyu Cai, David Wipf, Zuobai Zhang, and Zhaocheng Zhu for their helpful discussions and comments.
This project is supported by the Natural Sciences and Engineering Research Council (NSERC) Discovery Grant, the Canada CIFAR AI Chair Program, collaboration grants between Microsoft Research and Mila, Samsung Electronics Co., Ltd., Amazon Faculty Research Award, Tencent AI Lab Rhino-Bird Gift Fund and a NRC Collaborative R\&D Project (AI4D-CORE-06). This project was also partially funded by IVADO Fundamental Research Project grant PRF-2019-3583139727.
The Stanford team is supported by NSF(\#1651565, \#1522054, \#1733686), ONR (N000141912145), AFOSR (FA95501910024), ARO (W911NF-21-1-0125) and Sloan Fellowship.
\bibliography{reference}
\bibliographystyle{iclr2022_conference}

\newpage
\appendix
\section{Proofs}
\label{app:sec:proof}

\subsection{Properties of the Diffusion Model}

We include proofs for several key properties of the probabilistic diffusion model here to be self-contained. For more detailed discussions, please refer to \cite{ho2020denoising}.
Let $\{\beta_0,...,\beta_T\}$ be a sequence of variances, and $\alpha_t = 1 - \beta_t$ and $\bar{\alpha}_t = \prod_{s=1}^{t} \alpha_s$. The two following properties are crucial for deriving the final tractable objective in \eqref{eq:train-obj}.

\begin{property}
\label{app:property:q-marginal}
Tractable marginal of the forward process:
\begin{equation}
\begin{aligned}
q(\gC^t | \gC^0) &= \int q(\gC^{1:t}|\gC^0) \,d\gC^{1:{(t-1)}} = \gN(\gC^t;~\sqrt{\bar{\alpha}_t} \gC^0, (1-\bar{\alpha}_t) I) \nonumber .
\end{aligned}
\end{equation}
\end{property} 
\begin{proof}
Let $\epsilon_i$'s be independent standard Gaussian random variables. Then, by definition of the Markov kernels $q(\gC^{t}|\gC^{t-1})$ in \eqref{eq:diffusion}, we have
\begin{align}
\begin{array}{rl}
    \gC^t & =\sqrt{\alpha_t}\gC^{t-1}+\sqrt{\beta_t}\epsilon_t \\
     & = \sqrt{\alpha_t\alpha_{t-1}}\gC^{t-2}+\sqrt{\alpha_t\beta_{t-1}}\epsilon_{t-1}+\sqrt{\beta_t}\epsilon_t \\
     & = \sqrt{\alpha_t\alpha_{t-1}\alpha_{t-1}}\gC^{t-3}+\sqrt{\alpha_t\alpha_{t-1}\beta_{t-2}}\epsilon_{t-2}+\sqrt{\alpha_t\beta_{t-1}}\epsilon_{t-1}+\sqrt{\beta_t}\epsilon_t \\
     & = \cdots \\
     & = \sqrt{\bar{\alpha}_t}\gC^0 + \sqrt{\alpha_t\alpha_{t-1}\cdots\alpha_2\beta_1}\epsilon_1+\cdots+\sqrt{\alpha_t\beta_{t-1}}\epsilon_{t-1}+\sqrt{\beta_t}\epsilon_t 
\end{array}
\end{align}
Therefore $q(\gC^t|\gC^0)$ is still Gaussian, and the mean of $\gC^t$ is $\sqrt{\bar{\alpha}_t}\gC^0$, and the variance matrix is $(\alpha_t\alpha_{t-1}\cdots\alpha_2\beta_1+\cdots+\alpha_t\beta_{t-1}+\beta_t) I = (1-\bar{\alpha}_t) I$. Then we have:
\begin{align*}
    q(\gC^t|\gC^0)=\gN(\gC^t;~ \sqrt{\bar{\alpha}_t}\gC^0,~ (1-\bar{\alpha}_t)  I).
\end{align*}
This property provides convenient closed-form evaluation of $\gC^t$ knowing $\gC^0$:
\begin{equation*}
\label{eq:xt-from-x0}
    \gC^t = \sqrt{\bar{\alpha}_t} \gC^0 + \sqrt{1-\bar{\alpha}_t}\bm{\epsilon},
\end{equation*}
where $\bm{\epsilon} \sim \gN(0, I)$.

Besides, it is worth noting that,
\begin{align*}
    q(\gC^T|\gC^0)=\gN(\gC^T;~ \sqrt{\bar{\alpha}_T}\gC^0,~ (1-\bar{\alpha}_T) I),
\label{eq:q(xT|x0)}
\end{align*}
where $\bar{\alpha}_T = \prod_{t=1}^T (1 - \beta_t)$ approaches zero with large $T$, which indicates the diffusion process can finally converge into a whitened noisy distribution. 
\end{proof}

\begin{property}
\label{app:property:q-post}
Tractable posterior of the forward process:
\begin{equation*}
\label{eq:posterior-appendix}
    \begin{aligned}
    q(\gC^{t-1} | \gC^t, \gC^0) = \gN(\gC^{t-1};~\frac{\sqrt{\bar{\alpha}_{t-1}} \beta_t}{1-\bar{\alpha}_t} \gC^0 +\frac{\sqrt{\alpha_t} (1-\bar{\alpha}_{t-1})}{1-\bar{\alpha}_t} \gC^t,
    \frac{(1-\bar{\alpha}_{t-1})}{1-\bar{\alpha}_t}\beta_t I).
    \end{aligned}
\end{equation*}
\end{property}
\begin{proof}
    Let $\tilde{\beta}_t=\frac{1-\bar{\alpha}_{t-1}}{1-\bar{\alpha}_t}\beta_t$, then we can derive the posterior by Bayes rule:
    \begin{align}
    \begin{array}{rl}
        q(\gC^{t-1}|\gC^t,\gC^0) 
        & \displaystyle = \frac{q(\gC^t|\gC^{t-1})~q(\gC^{t-1}|\gC^0)}{q(\gC^t|\gC^0)} \\
        & \displaystyle = \frac{\gN(\gC^t;\sqrt{\alpha_t}\gC^{t-1},\beta_tI) ~\gN(\gC^{t-1};\sqrt{\bar{\alpha}_{t-1}}\gC^0,(1-\bar{\alpha}_{t-1})I)}{\gN(\gC^t;\sqrt{\bar{\alpha}_t}\gC^0,(1-\bar{\alpha}_t)I)} \\
        & \displaystyle = (2\pi\beta_t)^{-\frac d2}(2\pi(1-\bar{\alpha}_{t-1}))^{-\frac d2}(2\pi(1-\bar{\alpha}_t))^{\frac d2}\times \\
        & \displaystyle ~~~~ \exp\left(-\frac{\|\gC^t-\sqrt{\alpha_t}\gC^{t-1}\|^2}{2\beta_t}-\frac{\|\gC^{t-1}-\sqrt{\bar{\alpha}_{t-1}}\gC^0\|^2}{2(1-\bar{\alpha}_{t-1})}+\frac{\|\gC^t-\sqrt{\bar{\alpha}_t}\gC^0\|^2}{2(1-\bar{\alpha}_t)}\right) \\
        & \displaystyle = (2\pi\tilde{\beta}_t)^{-\frac d2}\exp\left(-\frac{1}{2\tilde{\beta}_t}\left\|\gC^{t-1}-\frac{\sqrt{\bar{\alpha}_{t-1}}\beta_t}{1-\bar{\alpha}_t}\gC^0-\frac{\sqrt{\alpha_t}(1-\bar{\alpha}_{t-1})}{1-\bar{\alpha}_t}\gC^t\right\|^2\right)
    \end{array}
    \end{align}
    Then we have the posterior $q(\gC^{t-1}|\gC^t,\gC^0)$ as the given form.
\end{proof}

\subsection{Proof of Proposition~\ref{prop:markov-equivariance}}

Let $T_g$ be some roto-translational transformations of a group element $g \in$ SE(3), and let $p(x_T)$ be a density which is SE(3)-invariant, \textit{i.e.}, $p(x_T)=p(T_g (x_T))$. If the Markov transitions $p(x_{t-1}|x_t)$ are SE(3)-equivariant, \textit{i.e.}, $p(x_{t-1}|x_t) = p(T_g (x_{t-1})|T_g (x_t))$, then we have that the density $p_{\theta}(x_{0})=\int p(x_T) p_{\theta}(x_{0: T-1}|x_T) \mathrm{d} \boldsymbol{x}_{1: T}$ is also SE(3)-invariant.

\begin{proof}
    \begin{equation}
    \begin{aligned}
        p_{\theta}(T_g(x_{0})) & = \int p(T_g(x_T)) p_{\theta}(T_g(x_{0: T-1})|T_g(x_T)) \mathrm{d} \boldsymbol{x}_{1: T} \\
        & = \int p(T_g(x_T)) \Pi_{t=1}^T p_{\theta}(T_g(x_{t-1})|T_g(x_t)) \mathrm{d} \boldsymbol{x}_{1: T}\\
        & = \int p(x_T) \Pi_{t=1}^T p_{\theta}(T_g(x_{t-1})|T_g(x_t)) \mathrm{d} \boldsymbol{x}_{1: T} \quad \text{(invariant prior $p(x_T)$)} \\
        & = \int p(x_T) \Pi_{t=1}^T p_{\theta}(x_{t-1}|x_t) \mathrm{d} \boldsymbol{x}_{1: T} \quad \text{(equivariant kernels $p(x_{t-1}|x_t)$)} \\
        & = \int p(x_T) p_{\theta}(x_{0: T-1}|x_T) \mathrm{d} \boldsymbol{x}_{1: T}\\
        & = p_{\theta}(x_{0})
    \end{aligned}
    \end{equation}
\end{proof}

\subsection{Proof of Proposition~\ref{prop:gfn-equivariant}}

In this section we prove that the output $\rvx$ of GFN defined in equation~\ref{eq:nfl-message}, \ref{eq:nfl-node} and \ref{eq:nfl-tensor} is translationally invariant and rotationally equivariant with the input $\gC$. Let $g \in \mathbb{R}^3$ denote any translation transformations and orthogonal matrices $R \in \mathbb{R}^{3 \times 3}$ denote any rotation transformations. let $R\rvx$ be shorthand for
$(R\rvx_1, \cdots, R\rvx_N)$.
Formally, we aim to prove that the model satisfies:
\begin{equation}
\label{app:eq:gfn-require}
    R\rvx^{l+1}, \rvh^{l+1}  = \mathrm{GFN}(R\rvx^l, R\gC + g, \rvh^l).
\end{equation}
This equation indicates that, given $\rvx^l$ already rotationally equivalent with $\gC$, and $\rvh^l$ already invariant, then such property can propagate through a single GFN layer to $\rvx^{l+1}$ and $\rvh^{l+1}$. 
\begin{proof}
Firstly, given that $\rvh^l$ already invariant to SE(3) transformations,
we have that the messages $\rvm_{ij}$ calculated from \eqref{eq:nfl-message} will also be invariant. This is because it sorely relies on the distance between two atoms, which are manifestly invariant to rotations $\|R\rvx_{i}^{l} - R\rvx_{j}^{l}\|^{2} = (\rvx_i^l- \rvx_j^l)^\top R^\top R (\rvx_{i}^{l} -\rvx_{j}^{l}) = (\rvx_i^l- \rvx_j^l)^\top I (\rvx_{i}^{l} -\rvx_{j}^{l}) = \|\rvx_{i}^{l} -\rvx_{j}^{l}\|^{2}$. Formally, the invariance of messages in \eqref{eq:nfl-message} can be written as:
\begin{equation}
    \rvm_{i, j} =\Phi_{m}\left(\rvh_{i}^{l}, \rvh_{j}^{l},\left\|R\rvx_{i}^{l}- R\rvx_{j}^{l}\right\|^{2}, e_{i j}\right) = \Phi_{m}\left(\rvh_{i}^{l}, \rvh_{j}^{l},\left\|\rvx_{i}^{l} - \rvx_{j}^{l} \right\|^{2}, e_{i j}\right).
\end{equation}
And similarly, the $\rvh^{t+1}$ updated from \eqref{eq:nfl-node} will also be invariant.

Next, we prove that the vector $\rvx$ updated from \eqref{eq:nfl-tensor} preserves rotational equivariance and translational invariance. 
Given $\rvm_{ij}$ already invariant as proven above, we have that:
\begin{align}
    \sum_{j \in \gN(i)} \frac{1}{d_{ij}}\left(R\rvc_{i} + g - R\rvc_{j} - g\right) \Phi_{x}\left(\rvm_{i, j}\right)
    &= R \sum_{j \in \gN(i)} \frac{1}{d_{ij}} \left(\rvc_{i} - \rvc_{j}\right) \Phi_{x}\left(\rvm_{i, j}\right)
    = R\rvx_{i}^{l+1}.
\end{align}
Therefore, we have that rotating and translating $\rvc$ results in the same rotation and no translation on $\rvx^{l+1}$ by updating through \eqref{eq:nfl-tensor}.

Thus we can conclude that the property defined in \eqref{app:eq:gfn-require} is satisfied.
\end{proof}

Having proved the equivariance property of a single GFN layer, then inductively, we can draw conclusion that a composition of $L$ GFN layers will also preserve the same equivariance.

\subsection{Proof of Proposition~\ref{prop:elbo}}

We first derive the variational lower bound (ELBO) objective in \eqref{eq:elbo}.
The ELBO can be calculated as follows:
\begin{align}
    \mathbb{E} \log p_{\theta}(\gC^0|\gG) 
    &=  \mathbb{E} \log\mathbb{E}_{q(\gC^{1:T}|\gC^0)} \Big[ \frac{p_{\theta}(\gC^{0:T-1}|\gG,\gC^T)\times p(\gC^T)}{q(\gC^{1:T}|\gC^0)} \Big] \nonumber \\
    &\geq \mathbb{E}_{q} \log \frac{p_{\theta}(\gC^{0:T-1}|\gG,\gC^T)\times p(\gC^T)}{q(\gC^{1:T}|\gC^0)} \nonumber\\
    & \displaystyle = \mathbb{E}_q\Big[\log p(\gC^T)-\sum_{t=1}^T\log\frac{p_{\theta}(\gC^{t-1}|\gG,\gC^t)}{q(\gC^t|\gC^{t-1})}\Big] \nonumber \\
    & \displaystyle= \mathbb{E}_q\Big[\log p(\gC^T)-\log\frac{p_{\theta}(\gC^0|\gG,\gC^1)}{q(\gC^1|\gC^0)}-\sum_{t=2}^T\Big(\log\frac{p_{\theta}(\gC^{t-1}|\gG,\gC^t)}{q(\gC^{t-1}|\gC^t,\gC^0)}+\log\frac{q(\gC^{t-1}|\gC^0)}{q(\gC^t|\gC^0)}\Big)\Big] \nonumber \\ 
    & \displaystyle = \mathbb{E}_q\Big[\log \frac{p(\gC^T)}{q(\gC^T|\gC^0)}-\log p_{\theta}(\gC^0|\gG,\gC^1)-\sum_{t=2}^T\log\frac{p_{\theta}(\gC^{t-1}|\gG,\gC^t)}{q(\gC^{t-1}|\gC^t,\gC^0)}\Big] \nonumber \\
    & \displaystyle = -\mathbb{E}_q\Big[ \kl{q(\gC^T|\gC^0)}{p(\gC^T)}+\sum_{t=2}^T \kl{q(\gC^{t-1}|\gC^t,\gC^0)}{p_{\theta}(\gC^{t-1}|\gG, \gC^t)}-\log p_{\theta}(\gC^0|\gG,\gC^1)\Big].
\end{align}
It can be noted that the first term $\kl{q(\gC^T|\gC^0)}{p(\gC^T)}$ is a constant, which can be omitted in the objective. 
Furthermore, for brevity, we also merge the final term $\log p_{\theta}(\gC^0|\gG,\gC^1)$ into the second term (sum over KL divergences), and finally derive that $\gL_\mathrm{ELBO}=\sum_{t = 1}^T
D_{\text{KL}}(q(\gC^{t-1}|\gC^t,\gC^0)\|p_\theta(\gC^{t-1}|\gG,\gC^t))$ as in \eqref{eq:elbo}. 

Now we consider how to compute the KL divergences as the proposition~\ref{prop:elbo}.
    Since both $q(\gC^{t-1}|\gC^t,\gC^0)$ and $p_{\theta}(\gC^{t-1}|\gG,\gC^t)$ are Gaussian share the same covariance matrix $\tilde{\beta}_tI$, the KL divergence between them can be calculated by the squared $\ell_2$ distance between their means weighed by a certain weights $\frac{1}{2\tilde{\beta}_t}$. By the expression of $q(\gC^t|\gC^0)$, we have the reparameterization that $\gC^t=\sqrt{\bar{\alpha}_t}\gC^0+\sqrt{1-\bar{\alpha}_t}\epsilon$. Then we can derive:
    \begin{align}
    \begin{array}{l}
        \mathbb{E}_q~ \kl{q(\gC^{t-1}|\gC^t,\gC^0)}{p_{\theta}(\gG,\gC^{t-1}|\gC^t)} \\
        \displaystyle = \frac{1}{2\tilde{\beta}_t}\mathbb{E}_{\gC^0} \left\|\frac{\sqrt{\bar{\alpha}_{t-1}}\beta_t}{1-\bar{\alpha}_t}\gC^0+\frac{\sqrt{\alpha_t}(1-\bar{\alpha}_{t-1})}{1-\bar{\alpha}_t}\gC^t-\frac{1}{\sqrt{\alpha_t}}\left(\gC^t-\frac{\beta_t}{\sqrt{1-\bar{\alpha}_t}}\epsilon_{\theta}(\gC^t,\gG,t)\right)\right\|^2 \\
        \displaystyle = \frac{1}{2\tilde{\beta}_t}\mathbb{E}_{\gC^0,\epsilon} \left\|\frac{\sqrt{\bar{\alpha}_{t-1}}\beta_t}{1-\bar{\alpha}_t}\cdot\frac{\gC^t-\sqrt{1-\bar{\alpha}_t}\epsilon}{\sqrt{\bar{\alpha}_t}}+\frac{\sqrt{\alpha_t}(1-\bar{\alpha}_{t-1})}{1-\bar{\alpha}_t}\gC^t-\frac{1}{\sqrt{\alpha_t}}\left(\gC^t-\frac{\beta_t}{\sqrt{1-\bar{\alpha}_t}}\epsilon_{\theta}(\gC^t,\gG,t)\right)\right\|^2 \\
        \displaystyle = \frac{1}{2\tilde{\beta}_t}\cdot\frac{\beta_t^2}{\alpha_t(1-\bar{\alpha}_t)}\mathbb{E}_{\gC^0,\epsilon} \left\|0\cdot \gC^t+\epsilon-\epsilon_{\theta}(\gC^t,\gG,t)\right\|^2 \\
        \displaystyle = \frac{\beta_t^2}{2\frac{1-\bar{\alpha}_{t-1}}{1-\bar{\alpha}_t}\beta_t\alpha_t(1-\bar{\alpha}_t)}\mathbb{E}_{\gC^0,\epsilon} \left\|\epsilon-\epsilon_{\theta}(\gC^t,\gG,t)\right\|^2 \\
        \displaystyle = \gamma_t \mathbb{E}_{\gC^0,\epsilon} \left\|\epsilon-\epsilon_{\theta}(\gC^t,t)\right\|^2,
    \end{array}
    \end{align}
    where $\gamma_t$ represent the wights $\frac{\beta_t}{2\alpha_t(1-\bar{\alpha}_{t-1})}$. And we finish the proof.

\subsection{Analysis of the invariant density in Sec.~\ref{subsec:equivariant-decoder}}

Given a geometric system $x \in \mathbb{R}^{N \cdot 3}$, we obtain the CoM-free $\hat{x}$ by subtracting its CoM.
This can be considered as a linear transformation:
\begin{equation}
    \hat{x} = Q x, \text{~~where~~} Q = I_3 \otimes \left(I_{N} - \tfrac{1}{N} \textbf{1}_{N} \textbf{1}_{N}^{T}\right) 
\end{equation}
where $I_{k}$ denotes the $k \times k$ identity matrix and $\textbf{1}_{k}$ denotes the $k$-dimensional vector filled with ones.
It can be noted that $Q$ is a symmetric projection operator, \textit{i.e.}, $Q^2 = Q$ and $Q^T = Q$. And we also have that $\operatorname{rank}[Q] = (N-1) \cdot 3$. Furthermore, let $U$ represent the space of CoM-free systems, we can easily have that $Qy = y$ for any $y \in U$ since the CoM of $y$ is already zero.

Formally, let $n=N \cdot 3$ and set $\mathbb{R}^{n}$ with an isotropic normal distribution $\rho = \mathcal{N}(0, I_n)$, then the CoM-free density can be formally written as $\hat \rho = \mathcal{N}(0, Q I_n Q^{T}) = \mathcal{N}(0, Q Q^{T})$.
Thus, sampling from $\hat \rho$ can be trivially achieved by sampling from $\rho$ and then projecting with $Q$. And $\hat \rho(y)$ can be calculated by $\rho(y)$ since for any $y \in U$ we have $\| y \|^{2}_{2} = \| Q y \|^{2}_{2}$, and thus $\rho(y) = \hat\rho(y)$.

And in this paper, with the SE(3)-equivariant Markov kernels of the reverse process, any CoM-free system will transit to another CoM-free system. And thus we can induce a well-defined Markov chain on the subspace spanned by $Q$. 

\section{Other related work}

\textbf{Protein structure generation.} There has also been many recent works working on protein structure folding. 
For example, Boltzmann generators \cite{noe2019boltzmann} use flow-based models to generate the structure of protein main chains. \cite{alquraishi2019protein} uses recurrent networks to model the amino acid sequences. \cite{ingraham2019protein} proposed neural networks to learn an energy simulator to infer the protein structures. 
Most recently, AlphaFold \cite{senior2020protein,jumper2021highly} has significantly improved the performance of protein structure generation.
Nevertheless, proteins are mainly linear backbone structures while general molecules are highly branched with various rings, making protein folding approaches unsuitable for our setting.

\textbf{Point cloud generation.} Recently, some other works \citep{Luo2021DiffusionPM,chibane2020implicit} has also been proposed for 3D structure generation with diffusion-based models, but focus on the point cloud problem. Unfortunately, in general, point clouds are not considered as graphs with various atom and bond information, and equivariance is also not widely considered, making these methods fundamentally different from our model.

\section{Experiment details}
\label{app:sec:train-details}

In this section, we introduce the details of our experiments. In practice, the means $\epsilon_\theta$ are parameterized as compositions of both typical invariant MPNNs~\citep{schutt2017schnet} and the proposed equivariant GFNs in Sec.~\ref{subsec:equivariant-decoder}.
As a default setup, the MPNNs for parameterizing the means $\epsilon_\theta$ are all implemented with $4$ layers, and the hidden embedding dimension is set as $128$. After the MPNNs, we can obtain the informative invariant atom embeddings, which we denote as $\rvh^0$. Then the embeddings $\rvh^0$ are fed into equivariant layers and updated with \eqref{eq:nfl-message}, \eqref{eq:nfl-node}, and \eqref{eq:nfl-tensor} to obtain the equivariant output.
For the training of \method, we train the model on a single Tesla V100 GPU with a learning rate of $0.001$ until convergence and Adam~\citep{kingma2013auto} as the optimizer. The practical training time is around 48 hours.
The other hyper-parameters of \method are summarized in Tab.~\ref{app:tab:hyperparameters}, including highest variance level $\beta_T$, lowest variance level $\beta_T$, the variance schedule, number of diffusion timesteps $T$, radius threshold for determining the neighbor of atoms $\tau$, batch size, and number of training iterations.

\begin{table}[htbp]
    \centering
    \caption{Additional hyperparameters of our \method.}
    \label{app:tab:hyperparameters}
        \begin{tabular}{c cccccccc}
        \toprule
        Task & $\beta_1$ & $\beta_T$ & $\beta$ scheduler & $T$ & $\tau$ & Batch Size & Train Iter.\\ 
        \midrule
        QM9 & 1e-7 & 2e-3 & $\operatorname{sigmoid}$ & 5000 & 10\AA & 64 & 1M \\
        Drugs & 1e-7 & 2e-3 & $\operatorname{sigmoid}$ & 5000 & 10\AA & 32 & 1M \\
        \bottomrule
        \end{tabular}
\end{table}

\section{Additional experiments}
\label{app:sec:exp}

\subsection{Results for GEOM-QM9}

The results on the GEOM-QM9 dataset are reported in Tab.~\ref{tab:qm9}.

\begin{table}[!t]
    \vspace{-12pt}
    \caption{Results on the \textbf{GEOM-QM9} dataset, without FF optimization.}
    \label{tab:qm9}
    \centering
    \resizebox{\textwidth}{!}{
    \begin{tabular}{l|cccc|cccc}
    \toprule[1.0pt]
     & \multicolumn{2}{c}{\shortstack[c]{COV-R (\%) $\uparrow$}} & \multicolumn{2}{c|}{\shortstack[c]{MAT-R (\si{\angstrom}) $\downarrow$ }}  & \multicolumn{2}{c}{\shortstack[c]{COV-P (\%) $\uparrow$}}  & \multicolumn{2}{c}{\shortstack[c]{MAT-P (\si{\angstrom}) $\downarrow$ }} \\
    Models & Mean & Median & Mean & Median & Mean & Median & Mean & Median \\
    \midrule[0.8pt]
    \CVGAE & 0.09 & 0.00 & 1.6713 & 1.6088 & - & - & - & - \\ 
    \GraphDG & 73.33 & 84.21 & 0.4245 & 0.3973 & 43.90 & 35.33 & 0.5809 & 0.5823 \\ 
    \CGCF & 78.05 & 82.48 & 0.4219 & 0.3900 & 36.49 & 33.57 & 0.6615 & 0.6427 \\
    \ConfVAE & 77.84 & 88.20 & 0.4154 & 0.3739 & 38.02 & 34.67 & 0.6215 & 0.6091 \\
    \GeoMol & 71.26 & 72.00 & 0.3731 & 0.3731 & - & - & - & - \\ 
    \ConfGF & 88.49 & 94.31 & 0.2673 & 0.2685 & 46.43 & 43.41 & 0.5224 & 0.5124 \\
    \midrule[0.3pt]
    \bf \method-A & \bf 90.54 & \bf 94.61 & 0.2104 & 0.2021 & 52.35 & 50.10 & 0.4539 & 0.4399 \\
    \bf \method-C & 90.07 & 93.39 & \bf 0.2090 & \bf 0.1988 & \bf 52.79 & \bf 50.29 & \bf 0.4448 & \bf 0.4267 \\
    \bottomrule[1.0pt]
    \end{tabular}
    }
\end{table}

\subsection{Ablation study with fewer diffusion steps}

We also test our method with fewer diffusion steps. Specifically, we test the setting with $T=1000$, $\beta_1 = $1e-7 and $\beta_T=$9e-3. The results on the more challenging Drugs dataset are shown in Tab.~\ref{tab:fewer-decoding}. Compared with the results in Tab.~\ref{tab:drugs}, we can observe that when setting the diffusion steps as 1000, though slightly weaker than the performance with 5000 decoding steps, the model can already outperforms all existing baselines. Note that, the most competitive baseline \ConfGF~\citep{shi*2021confgf} also requires 5000 sampling steps, which indicates that our model can achieve better performance with fewer computational costs compared with the state-of-the-art method.

\begin{table}[t]
    \centering
    \small
    \caption{Additional results on the \textbf{GEOM-Drugs} dataset, without FF optimization.}
    \label{tab:fewer-decoding}
    \resizebox{\linewidth}{!}{
    \begin{tabular}{l|cccc|cccc}
    \toprule[1.0pt]
     & \multicolumn{2}{c}{\shortstack[c]{COV-R (\%) $\uparrow$}} & \multicolumn{2}{c|}{\shortstack[c]{MAT-R (\si{\angstrom}) $\downarrow$ }}  & \multicolumn{2}{c}{\shortstack[c]{COV-P (\%) $\uparrow$}}  & \multicolumn{2}{c}{\shortstack[c]{MAT-P (\si{\angstrom}) $\downarrow$ }} \\
    Models & Mean & Median & Mean & Median & Mean & Median & Mean & Median \\
    \midrule[0.8pt]
    \textbf{\method (T=1000)} & 82.96 & 96.29 & 0.9525 & 0.9334 & 48.27 & 46.03 & 1.3205 & 1.2724 \\
    \bottomrule[1.0pt]
    \end{tabular}
    }
    \vspace{-5pt}
\end{table}

\section{More Visualizations}
\label{app:sec:sample}

\begin{figure}[!ht]
	\centering
    \includegraphics[width=\linewidth]{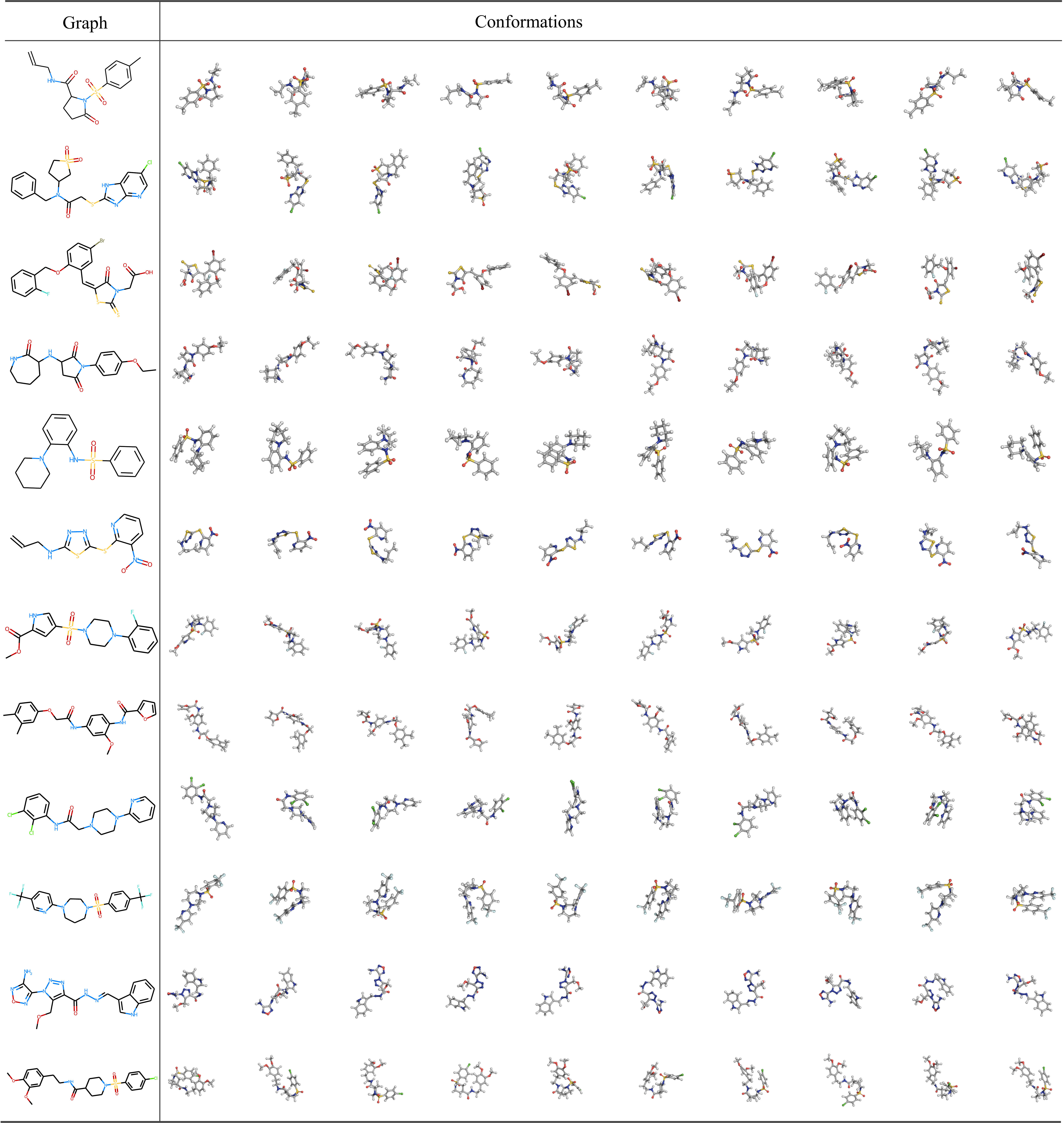}
    \vspace{-12pt}
    \caption{Visualization of drug-like conformations generated by \method.}
    \label{fig:exp_suppl}
    \vspace{-12pt}    
\end{figure}

We provide more visualization of generated structures in Fig.~\ref{fig:exp_suppl}. The molecules are chosen from the test split of GEOM-Drugs dataset.

\end{document}